# A Comparison of Stereo-Matching Cost between Convolutional Neural Network and Census for Satellite Images

**Bihe Chen** [b], Ph.D. Student
**Rongjun Qin** [a,b], Assistant Professor
**Xu Huang** [a], Postdoctoral Fellow
**Shuang Song** [a], Ph.D. Student
**Xiaohu Lu** [a], Ph.D. Student

[a] Department of Civil, Environmental and Geodetic Engineering, The Ohio State University,
218B Bolz Hall, 2036 Neil Avenue, Columbus, OH 43210, USA;
[b] Department of Electrical and Computer Engineering, The Ohio State University,
205 Dreese Labs, 2015 Neil Avenue, Columbus, OH 43210, USA

chen.8868@osu.edu
qin.324@osu.edu
huang.3651@osu.edu
song.1634@osu.edu
lu.2037@osu.edu

**ABSTRACT**

Stereo dense image matching can be categorized to low-level feature based matching and deep feature based matching according to their matching cost metrics. Census has been proofed to be one of the most efficient low-level feature based matching methods, while fast Convolutional Neural Network (fst-CNN), as a deep feature based method, has small computing time and is robust for satellite images. Thus, a comparison between fst-CNN and census is critical for further studies in stereo dense image matching. This paper used cost function of fst-CNN and census to do stereo matching, then utilized semi-global matching method to obtain optimized disparity images. Those images are used to produce digital surface model to compare with ground truth points. It addresses that fst-CNN performs better than census in the aspect of absolute matching accuracy, histogram of error distribution and matching completeness, but these two algorithms still performs in the same order of magnitude.

**KEYWORDS:** stereo matching, matching cost, convolutional neural network, census

## 1. INTRODUCTION

Stereo dense image matching is a critical component in generating 3D point clouds for mapping. Such methods highly rely on computing the appearance similarity of corresponding pixels in a pair of images, which is normally called matching cost. Each matching cost metric has its own way to extract the appearance features for each pixel. Most matching cost metrics only extract the low-level features (intensities, gradients, structures, etc.) for matching (Hirschmuller and Scharstein, 2009), while some metrics extract the deep features from convolutional neural networks (Lecun et al. 1998), therefore these matching cost metrics can be categorized into 1) low-level feature based matching cost; 2) deep feature based matching cost.

Low-level feature based matching cost, such as sum of absolute intensity differences (SAD) (Hirschmuller and Scharstein, 2007), normalized cross correlation (NCC) (Scharstein and Richard, 2001), mutual information (MI) (Hirschmuller and Scharstein, 2007)
 and census (Hirschmuller and Scharstein, 2007), computes the appearance similarities of correspondences by comparing their low-level features (intensities, gradients structures, etc.). These low-level feature based matching costs are superior in low time complexity and high flexibility in variable datasets (e.g. satellite images, Aerial images, street-view images and indoor images). However, it is generally known that these cost metrics are sensitive to geometric distortion, radiometric differences, textures (especially untextured regions or repeated texture regions), etc. between image pairs. The resulting uncertainties in matching cost often significantly reduce the accuracies of the final matching point clouds.

Deep feature based matching cost was proposed in recent years (Zbontar and LeCun, 2015), which compared appearances similarity of image patches by training a convolutional neural network (CNN). The training datasets are a large quantity (tens of millions) of image patch pairs with true disparities (e.g. obtained by LiDAR or structured light). The radiometric and geometric information of some of the image patch pairs were also adjusted such that the trained model is able to compute accurate matching cost in geometric distortion or radiometric difference scenarios. As deep feature is able to describe the appearance similarity more accurately than low-level features, most state of the art stereo matching methods utilized the trained CNN model to obtain high ranks in Middlebury and KITTI benchmark. However, the training datasets and testing datasets for the CNN model are in the same scenario. For example, in KITTI benchmark, the CNN model is trained by KITTI street-view training datasets, and also tested by KITTI street-view testing datasets. Critical analysis has not yet been done when training datasets and testing datasets are in different scenarios. For example, the training sets are indoor or street view images, while the testing datasets are satellite images. Considering the large geometric and radiometric differences among satellite images, aerial images, street-view images and indoor images, it needs to test and analysis the flexibility of deep feature based matching cost in different imaging scenarios.

To give a more widely evaluation of the deep feature based matching cost, we performed comparisons of the deep feature based matching cost and low-level feature based matching cost on different image datasets with different radiometric conditions. We chose census as the low-level feature based cost in the comparisons, which has been proven to be one of the most effective and radiation-invariance way for cost computation (Hirschmuller and Scharstein, 2009), and chose fast CNN model (fst-CNN) (Girshick, 2015) trained by KITTI benchmark as the deep feature based cost, which is suitable for matching in satellite image datasets, considering its low time complexity. We respectively selected three satellite image pairs in four typical regions in Argentina for comparisons: 1) small building region, which aimed at testing performances of both metrics in fine structures; 2) tall building region, which compared their performances in large geometric distortion scenario; 3) medium building region, which tested their matching reliability in normal areas. We manually enlarged radiometric differences in some of the pairs and respectively used census and fst-CNN model to compute matching cost, and then utilized semi-global matching method (SGM) to obtain optimized disparity images. The performances of them were comprehensively evaluated in four metrics: 1) absolute matching accuracies, which is computed with ground truth points (from LiDAR); 2) Histogram of error distribution; 3) Existence of systematic errors, which checks if systematic error exists. 4) Matching completeness, which measures noises in the final matching results.

The rest of the paper is organized as follows: Section 2 simply introduces the principles of census and fst-CNN model; Section 3 describes our evaluation metrics for testing census and fst-CNN; Section 4 shows the experimental results; and Section 5 draws the conclusions based on our works.

## 2. MATCHING COSTS COMPUTATION

Though promising in dense matching of indoor and street-view image datasets, deep feature based matching cost (e.g. fst-CNN) has not proved its ability in satellite images, among which fst-CNN has a linear time complexity $O(m \cdot n)$ with $n$ being the number of pixels and $m$ being the computation times for each pixel. Considering the large scales of satellites, we choose fst-CNN for the comparison. We also choose census as representative of low-level based matching cost, which has achieved great success in dense matching of satellite images (Qin, 2017). No matter fst-CNN or census, both of their principles are to compare similarity between image patches. In epipolar pairs, their principles can be concluded as:

$$C(\bm{p}, d) = F(w_I(\bm{p}), w_{I'}(\bm{p'}, d)) \tag{1}$$

where, $I$, $I'$ are a pair of images with $I$ being a basic image (normally the left image) and $I'$ being the matched image (normally the right); $\bm{p}$ is a pixel in the basic image; $\bm{p'}$ is the corresponding pixel in the matched image; $d$ is a disparity which measures the column coordinate difference between correspondence, $d = x_p - x_{p'}$, with $x_p$, $x_{p'}$ being the column coordinates of $\bm{p}$ and $\bm{p'}$; $w_I(\bm{p})$ is a basic image patch centered at $p$; $w_{I'}(\bm{p'}, d)$ is a matched image patch centered at the $\bm{p'}$; $C(\bm{p}, d)$ is the matching cost of $\bm{p}$ with corresponding disparity $d$, smaller cost means more similar image patches or vice versa. $F$ is a similarity computation function between image patches. For census, $F$ is a binary coding function, and for fst-CNN, $F$ is a convolutional neural network

**1. Census cost metric**

Census encodes local image structures in binary string with relative orderings of the pixel intensities, thus tolerating radiometric changes between image patches. It firstly compares intensities of neighbor pixels with that of the center pixel, and assigns 0 for pixels with larger intensities, and 1 for pixels with smaller intensities, thus giving a binary string for each image patch:

$$S(\boldsymbol{p}) = \{T(\boldsymbol{p}, \boldsymbol{q}) | \boldsymbol{q} \in w_I(\boldsymbol{p})\} \qquad (2)$$

where, $\boldsymbol{q}$ is a pixel in the image patch centered at pixel $\boldsymbol{p}$; $T(\boldsymbol{p}, \boldsymbol{q})$ is an indicator function, which is 1 if the intensity of $\boldsymbol{p}$ is large than $\boldsymbol{q}$, and 0, otherwise; $S(\boldsymbol{p})$ is a binary string which describes the local structure of $\boldsymbol{p}$.

The binary strings are computed for both basic image patch and matched image patch, and then a hamming distance metric is used to compare the similarity of the two strings:

$$F_{Census}(\boldsymbol{p}, d) = ||S(w_I(\boldsymbol{p})) - S(w_{I'}(\boldsymbol{p}', d))||_H \qquad (3)$$

where, $||\cdot||_H$ is hanming distance metric which returns the number of positions at which the corresponding symbols are different; $F_{Census}$ is the similarity function for census cost metric.

**2. fst-CNN based cost metric**

Fst-CNN (Girshick, 2015) is a network that measures the similarity of two input patches. The weights of two sub-networks are tied. The sub-networks are composed of a number of four convolutional layers with a ReLU activation function except the last layer. Both sub-networks output a feature vector extracted from the input patches. After normalization, the resulting two vectors are compared using dot product to produce the cosine similarity of the features extracted from the input patch pairs. An overview of the architecture is shown in Figure 1.

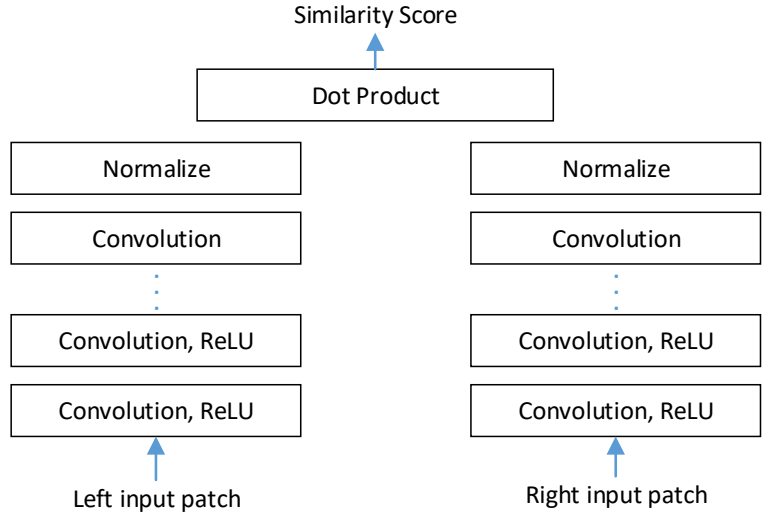

Figure 1. The architecture of fst-CNN (Girshick, 2015).

The inputs are two image patches and the output is the similarity score between 0 and 1.

The network is trained by minimizing a hinge loss. The loss is computed by considering pairs of examples centered around the ground truth position where one example belongs to the positive and one to the negative class. Then, then trained network can be utilized to compute the matching cost:

$$F_{fst-CNN}(\boldsymbol{p}, d) = -V(w_I(\boldsymbol{p})) \cdot V(w_{I'}(\boldsymbol{p}', d)) \qquad (4)$$

where, $w_I(\boldsymbol{p})$, $w_{I'}(\boldsymbol{p}', d)$ are two input patches; $V$ is a feature vector from the sub-networks; $F_{fst-CNN}$ is the similarity function for fst-CNN cost metric.

## 3. EVALUATION METRICS

To comprehensively compare census and fst-CNN on different satellite image datasets, we need several matching accuracy metrics to evaluate their performances. We firstly derived digital surface model (DSM) from disparity images, and then evaluate performances of census and fst-CNN by comparing the geometric accuracy of these DSMs. The most intuitive way is to measure the difference between the DSMs and the ground truth, which is called absolute matching accuracies. However, sometimes the high absolute accuracies do not mean good matching results. There may be still several tiny noises (below one pixel) on surfaces of DSM, reducing its surface smoothness. Therefore, we need a metric to evaluate the surface smoothness of DSMs, also defined as relative matching accuracies. Matching robustness is another factor under consideration, especially in weak textured regions. The metric to measure matching robustness is called matching confidence. Finally, we need to detect and remove matching outliers, and count the percentage of matching inliers, which is called matching completeness.

**1) Absolute matching accuracy**

Given ground truth (normally from LiDAR), we can compare the DSMs with it by measuring their differences. In this paper, we used average of absolute disparity differences as the absolute matching accuracy metric:

$$M_{ab} = \frac{1}{|I|} \sum_{p \in I} \min(|d(\boldsymbol{p}) - d_g(\boldsymbol{p})|, \sigma) \tag{5}$$

where, $\boldsymbol{p}$ is a point in DSM $I$; $|I|$ is the number of points in $I$; $d(\boldsymbol{p})$ is the estimated elevation of $\boldsymbol{p}$; $d_g(\boldsymbol{p})$ is the corresponding ground truth; $\sigma$ is the upper bound of maximum elevation difference, which reduces the influence of extremely large noises on the final accuracy evaluation; $M_{ab}$ is the metric to evaluate absolute matching accuracy.

**2) Histogram of error distribution**

We used histogram to visualize the distribution of absolute matching accuracy. We chose error values between 0 and σ with a bin size of 0.1 meters and calculate the number of points lying in each bin. Then draw the histogram, let error be x-axis and number of points be y-axis.

**3) Existence of systematic errors**

Existence of systematic errors measures whether systematic error exists or not. If an evaluating system contains only random errors, then its results can be trusted, while systematic error may lead to incorrect statics, thus affect results. We define systematic error as the minus average of disparity differences:

$$M_{sys} = \frac{1}{|I|} \sum_{p \in I} sign(d(\boldsymbol{p}) - d_g(\boldsymbol{p})) * min(|d(\boldsymbol{p}) - d_g(\boldsymbol{p})|, \sigma) \tag{6}$$

where, $sign(d(\boldsymbol{p}) - d_g(\boldsymbol{p}))$ is sign function.

**4) Matching completeness**

Matching completeness is another metric to measure the matching robustness between image pairs. A good metric should make the matching results in left and right images consistent, otherwise, the inconsistent matching results will be identified as outliers, and remove them away, leaving invalid regions in DSMs or disparity images. In this paper, we compare the invalid point numbers and valid point numbers, and compute the matching completeness:

$$M_{cpl} = 1 - N_{invalid}/N_{valid} \tag{7}$$

where, $N_{invalid}$ is the number of invalid points, and $N_{valid}$ is the number of valid points; $M_{cpl}$ is the metric to evaluate matching completeness.

## 4. EXPERIMENTAL COMPARISION AND ANALYSIS

We selected 5 world-view 3 image pairs in Argentina regions for the experimental comparisons of Census and fst-CNN. The collected data of these images very from Marth 15$^{th}$ 2015 to December 18$^{st}$ 2015 with ground sampling distance (GSD) 0.3 m. To give a comprehensive comparison between Census and fst-CNN, we selected three typical testing regions: small buildings (Figure 2(a)), tall buildings (Figure 2(b)) and repetitive-structure buildings (Figure 2(c)). The small building regions tend to have similar disparities, thus testing the performances of both cost metric in such fine structures. The tall building regions often have large disparity changes which will bring some matching uncertainties as well as occlusions. The repetitive-structure building region has repetitive textures which will bring high matching uncertainties, thus reducing matching accuracies.

For each pair, we respectively used Census and fst-CNN to compute cost volumes, then used semi-global matching (SGM) method to compute the corresponding disparity image, and transformed the disparity image into DSM using orientation parameters. After then, we fused all pairs of DSMs together by median filtering (Figure 3), compared and analyzed these DSMs using the four matching accuracy metric in section 3. In all matching accuracy evaluations, we force the maximum matching errors being 10 m, thus the matching accuracy evaluation will not be influenced by few extremely large errors.

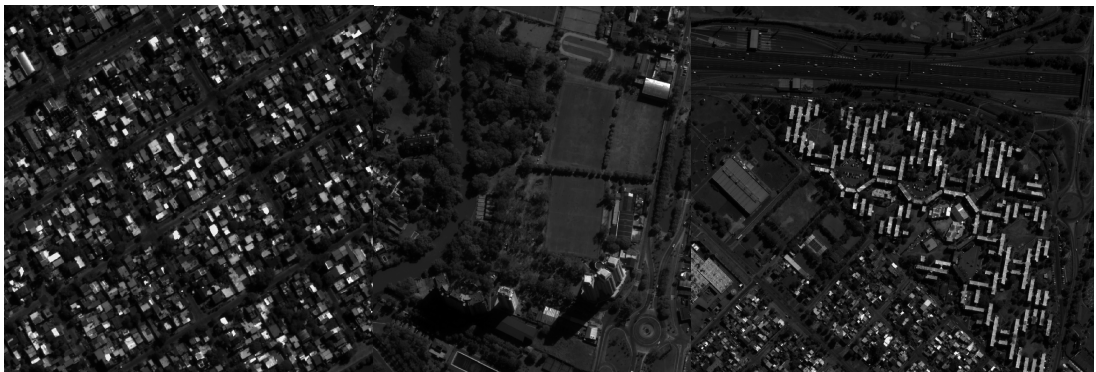

| 2(a) small buildings | 2(b) tall buildings | 2(c) medium buildings |

Figure 2. Images of Testing Regions

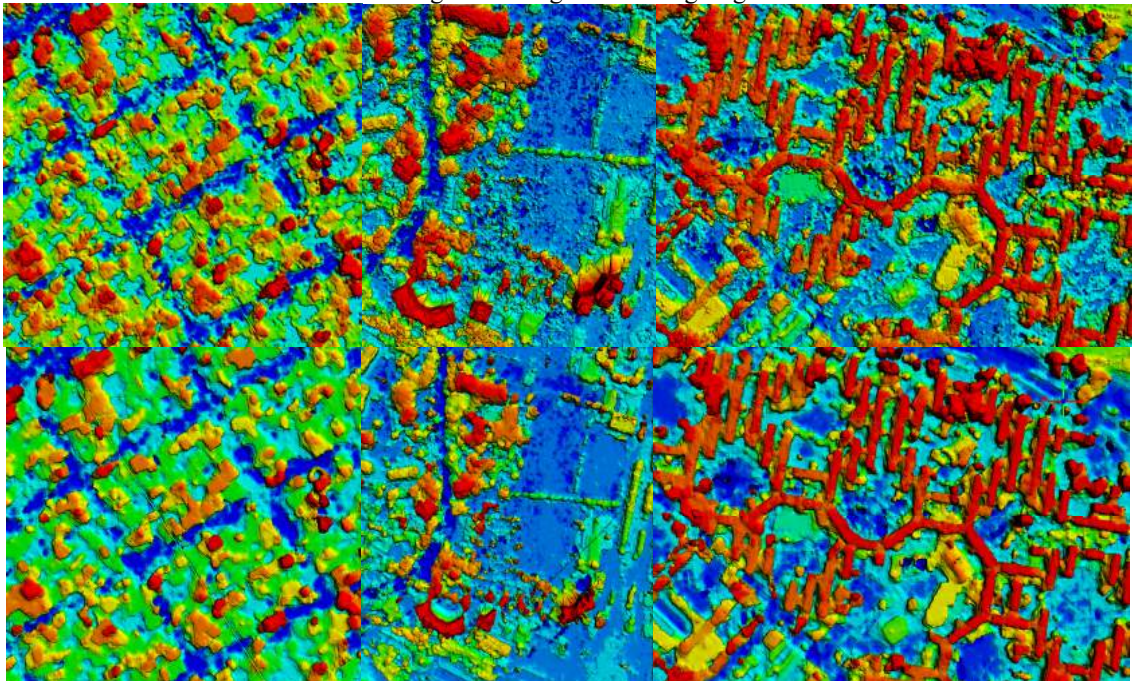

| 3(a) small buildings | 3(b) tall buildings | 3(c) medium buildings |

Figure 3. Resulting DSM

Figure 3 shows that the fst-CNN is capable of computing more robust and more accurate cost volume than census cost with less noises and smoother surfaces. The absolute accuracies of fst-CNN in three regions were 1.4416, 2.2898 and 1.6028, while the accuracies of census were only 1.4869, 2.5351 and 1.8674, which showed that fst-CNN is superior to census in different types of regions. We also checked the error distributions of census and fst-CNN in histogram, as shown in Figure 4.

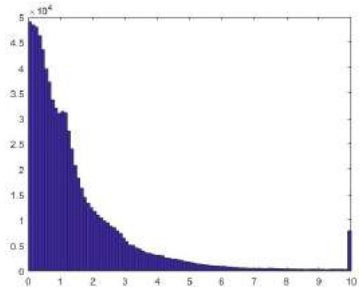 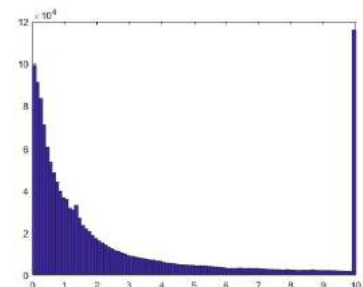 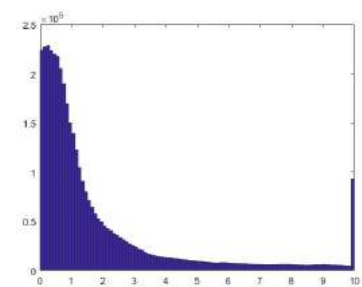

| 4(a) small buildings (census) | 4(b) tall buildings (census) | 4(c) medium buildings (census) |

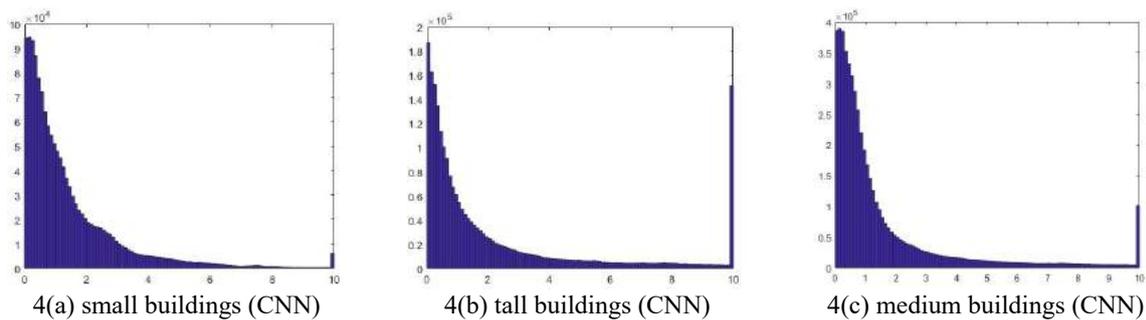

| 4(a) small buildings (CNN) | 4(b) tall buildings (CNN) | 4(c) medium buildings (CNN) |

Figure 4 Histogram of Error Distribution

The error distributions of fst-CNN and census were similar. As the maximum errors have been forced being 10 m, there was a small peak at the end the histograms. The length of the last bin of fst-CNN is shorter than census, which shows that fst-CNN has less matching outliers, and the length of the first bin of fst-CNN is longer than census, which shows that fst-CNN has more accurate points. The reason fst-CNN is more accurate than census is that, first, fst-CNN can extract and make use of deep features while Census simply uses low features, i.e. gray value. This makes fst-CNN more robust than census for it has more valuable information. Then, during the training process, part of the training data's radiometric conditions has been manually adjusted. This makes fst-CNN more robust in dealing with radiometric distortions than Census. Last, geometric distortions were also added to some of the data in the training process, and this makes fst-CNN more robust to geometric distortions.

We also checked if both cost metric contained systematic bias in matching. Assuming that the expectation of all matching accidental errors is zero, the systematic errors can be computed by averaging all matching errors. The averaging results are the corresponding systematic errors. Following Equation 6, the systematic errors of fst-CNN and census were shown in Table 1.

|  | CNN | RSP |
|---|---|---|
| Small building | -0.0400 | 0.1167 |
| Tall building | -0.4103 | -0.2767 |
| Explorer region | -0.3125 | -0.2639 |

Table 1. comparison of systematic errors

All system errors of both metrics in Table 1 were close to 0, thus indicating that both metrics do not contain systematic errors.

Finally, we tested the matching completeness of both metrics, as shown in Table 2.

|  | CNN | RSP |
|---|---|---|
| Small building | **0.9816** | 0.9507 |
| Tall building | **0.9163** | 0.8347 |
| Explorer region | **0.9648** | 0.9140 |

Table 2.

Table 2 shows fst-CNN is able to achieve more complete matching results than census, especially in areas with tall buildings. This is because CNN makes use of deep features thus performs better in large geometric distortion scenarios.

## 5. CONCLUSION

In this paper, we compared stereo matching results between census and fst-CNN in different perspectives and showed that fst-CNN performs better in all three metrics: 1) absolute matching accuracies; 2) Histogram of error distribution; 3) Matching completeness. Also, we showed that CNN performs in the same order of magnitude with census, which matched intuitive inference.

In our future work, we plan to compare low-level feature based matching methods and deep feature based matching using images with different convergence angles.

## ACKNOWLEDGEMENT

The study is partially supported by the ONR grant (Award No. N000141712928). The author would like to thank John Hopkins University Applied Physics Lab to support the Imagery, and the author would also like to thank the IARPA to organize the 3D challenge available that drives forth this work.